\documentclass[twoside]{article}
\usepackage{graphicx}
\usepackage{multirow}
\usepackage{url}
\usepackage{amssymb}
\usepackage{hyperref} 
\usepackage[accepted]{aistats2026}
\usepackage[round]{natbib}

\begin{document}

%

%

\twocolumn[

\aistatstitle{From Hawkes Processes to Attention: Time-Modulated Mechanisms for Event Sequences}

\aistatsauthor{ Xinzi Tan \And Kejian Zhang \And Junhan Yu \And Doudou Zhou$^*$}

\aistatsaddress{ Department of Statistics and Data Science, National University of Singapore } 
\vspace{-2\baselineskip}
\begin{center}
\texttt{\{tan.xinzi, kejianzhang, junhan.yu\}@u.nus.edu, ddzhou@nus.edu.sg}\\
\small $^*$ Corresponding author
\end{center}
\vspace{0.25\baselineskip}
]

\begin{abstract}
Marked Temporal Point Processes (MTPPs) arise naturally in medical, social, commercial, and financial domains. However, existing Transformer-based methods mostly inject temporal information only via positional encodings, relying on shared or parametric decay structures, which limits their ability to capture heterogeneous and type-specific temporal effects. Inspired by this observation, we derive a novel attention operator called \textit{Hawkes Attention} from the multivariate Hawkes process theory for MTPP, using learnable per-type neural kernels to modulate query, key and value projections, thereby replacing the corresponding parts in the traditional attention. Benefited from the design, Hawkes Attention unifies event timing and content interaction, learning both the time-relevant behavior and type-specific excitation patterns from the data. The experimental results show that our method achieves better performance compared to the baselines. In addition to the general MTPP, our attention mechanism can also be easily applied to specific temporal structures, such as time series forecasting.
\end{abstract}

\section{INTRODUCTION}
Marked Temporal Point Processes (MTPPs) provide a principled framework for modeling event sequences in continuous time. They arise in diverse domains, from electronic health records \citep{wang2018supervised}, where patient trajectories are captured as diagnoses, tests, and treatments, to social networks \citep{yang2011like}, where interactions unfold with irregular timing, and to high-frequency finance \citep{bacry2015hawkes}, where transaction intervals drive volatility and risk. In these settings, predictive models must jointly capture \emph{what} that occurs (type or value of the event), and \emph{when} that occurs (timestamp or interval).

Classical models, such as the Poisson and multivariate Hawkes processes \citep{daley2008introduction,liniger2009multivariate}, capture temporal excitation via hand-crafted kernels (e.g., exponential decay). While interpretable, these kernels impose rigid functional forms and rely on linear superposition, which limits their ability to model heterogeneous or higher-order interactions \citep{mei2017neural}.

A series of neural models replaces hand-crafted Hawkes kernels with learned dynamics. RMTPP \citep{du2016recurrent} uses recurrent networks to model event times and types jointly, while NHP \citep{mei2017neural} employs a continuous-time LSTM to modulate intensity. FullyNN \citep{omi2019fullynn} parameterizes intensities end-to-end with neural networks, and IFTPP \citep{shchur2020intensityfree} bypasses intensity functions via transport maps. ODETPP \citep{chen2021neural} integrates neural ordinary differential equations for spatio-temporal dynamics. Although powerful, these models inherit sequential inefficiency, struggle with long-range dependencies, and lack decomposable, interpretable influence kernels.

Self-attention has been adapted to MTPPs to address these issues. THP \citep{zuo2020transformer} applies Transformer-style attention to continuous-time events, while SAHP \citep{zhang2020self} uses accumulated attention weights to quantify influences. AttNHP \citep{yang2022transformer} combines attention with neural Hawkes dynamics. These methods improve scalability and capture long-range dependencies, but typically handle time only indirectly, either through positional encodings or fixed decay terms, which limits both their flexibility and interpretability.

Beyond neural and attention-based MTPPs, several works extend temporal point processes (TPPs) with additional structure. Recent studies have adapted Transformers and neural ordinary differential equations for irregular time series \citep{irani2024positional}, highlighting the importance of directly encoding elapsed time. NJDTPP \citep{zhang2024neural} models TPP intensities as neural jump-diffusion stochastic differential equations and \citet{bae2023meta} frames TPPs as neural processes for meta-learning, using context sets and attentive local history matching for fast adaptation. PromptTPP \citep{NEURIPS2023_3c129892} brings continual learning to neural TPPs by maintaining a continuous-time prompt pool. Spatial-temporal point process models \citep{NEURIPS2023_4eb2c0ad} incorporate location along with time, while domain-specific variants have been proposed in epidemiology, finance, and healthcare \citep{zhou2013learning,pang2021cehr}. These extensions underscore the broad applicability of point process models, but have not resolved the previous issues. 

To overcome these challenges, we propose \textit{Hawkes Attention}, a new attention mechanism derived directly from the mathematical structure of Hawkes processes. Our operator embeds the elapsed time in the core of query, key, and value computations, removing the need for positional encodings. Instead of fixed parametric kernels, we learn flexible per-type neural influence functions that capture heterogeneous patterns while preserving interpretability, separating influence magnitude and temporal profile. Through low-rank factorization of pairwise coefficients, Hawkes Attention balances expressivity with efficiency, yielding a time-aware attention operator that generalizes Hawkes processes while retaining scalability and parallelism.

Our contributions are:
\begin{itemize}
    \item A principled derivation of a time-modulated attention operator from multivariate Hawkes process intensities, bridging probabilistic point process modeling with modern attention mechanisms.
    \item Per-type neural kernels that flexibly capture heterogeneous temporal influences while preserving interpretability.
    \item A fully time-aware attention mechanism that eliminates positional encodings and extends naturally to both irregular MTPPs and regularly sampled time series.
\end{itemize}

The remainder of this paper is organized as follows: {Section~\ref{sec:methods}} presents the mathematical derivation and architectural details of Hawkes Attention. {Section~\ref{sec:exp}} describes our experimental protocols, datasets, and results. {Section~\ref{sec:dis_conclu}} first discusses the generalization of our method to traditional time series, providing a general framework for modeling sequential data, and finally concludes our work. Our code is available on Github\footnote{https://github.com/TanXZfra/Hawkes-Attention}.

\section{FROM HAWKES PROCESS TO ATTENTION}
\label{sec:methods}

We begin from the classical Hawkes formulation and progressively introduce three modifications that yield our Hawkes Attention model: (i) a low-rank decomposition of the influence matrix into event embeddings, (ii) flexible per-type neural kernels for temporal profiles, and (iii) an attention-based interactive aggregator in place of linear superposition. Each step addresses a specific limitation of Hawkes processes and naturally leads to the final architecture.

\subsection{Hawkes process decomposition}
We first model asynchronous event sequences as 
\begin{equation}
\mathcal{H} = \{ (t_1, c_1), (t_2, c_2), \dots, (t_m, c_m) \},
\label{eq:event_sequence}
\end{equation}
where $m$ is the length of the sequence, $t_k$ is the time of occurrence of the $k$-th event with $t_1 \leq t_2 \leq \dots \leq t_m$; $c_k \in \mathcal{C}$ is the type of the $k$-th event with $\mathcal{C}$ being the set of all distinct types of events and $|\mathcal{C}|$ being the cardinality of $\mathcal{C}$. MTPPs characterize the likelihood that the new event at timestamp $t$ is type-$c$ through the type-specific conditional intensity:
\begin{equation}
\lambda_c(t \mid \mathcal{H}_t) = \lim_{\Delta t \to 0} 
\frac{\mathbb{E}\bigl[ \#\{c\text{-events in } [t, t+\Delta t)\} \mid \mathcal{H}_t \bigr]}{\Delta t},
\label{eq:mtpp_intensity}
\end{equation}
with $\mathcal{H}_t=\{(t_k,c_k):t_k<t\}$ the history of events.

In a multivariate Hawkes process \citep{liniger2009multivariate}, the conditional intensity of type-$c$ event at timestamp $t$ is
\begin{equation}
\lambda_c(t|\mathcal{H}_t)
= \mu_c + \sum_{(t_k,c_k)\in\mathcal{H}_t}\Phi_{c,c_k}(t-t_k),
\label{eq:hawkes}
\end{equation}
where $\mu_c$ is the base intensity of type $c$ and $\Phi_{c,c_k}$ is a kernel that describes how a past event of type $c_k$ influences the likelihood of type $c$ over the elapsed time $\Delta t=t-t_k$.

Although the multivariate Hawkes formulation in Eq.~\eqref{eq:hawkes} provides clear probabilistic semantics, it is practically infeasible to learn a distinct, unconstrained kernel $\Phi_{c,c_k}$ for every ordered pair $(c,c_k)$ when $|\mathcal C|$ is large. To obtain a tractable yet expressive parameterization, we factorize the pairwise kernel into a type-specific temporal profile and a target-dependent magnitude:
\begin{equation}
\Phi_{c, c_k}(t - t_k) = \beta_{c, c_k} \cdot \phi_{c_k}(t - t_k),
\label{eq:kernel_decomposition}
\end{equation}
where $\phi_{c_k}(\cdot)$ is a fixed temporal decay (e.g.\ exponential) describing the temporal profile depending on the type of event $c_k$, and $\beta_{c,c_k}$ are learnable coefficients representing static magnitudes that describe how strongly the type $c_k$ influences $c$. This separation aligns with Hawkes' intuition and drastically reduces the number of distinct temporal functions to learn, while allowing for more flexible modeling of $\phi_{c_k}$ and providing greater interpretability. In practice, this simplifies visual interpretation and aggregation across heads and types.

To express the intensity in vectorized form under this parameterization, we group the magnitude terms by the type of the target event $c$ into an influence vector $\boldsymbol{\beta}_c = (\beta_{c,1}, \dots, \beta_{c, |\mathcal{C}|})^\top$. The total influence can then be viewed as an inner product between this vector and the sum of a series of corresponding vectors $\boldsymbol\phi_{c_k}(t - t_k)$, where each $\boldsymbol\phi_{c_k}(t - t_k)$ is a sparse vector that has a nonzero value $\phi_{c_k}(t-t_k)$ only at the $c_k$-th index. The intensity function can thus be rewritten as:
\begin{equation}
\lambda_c(t | \mathcal{H}_t) = \mu_c + \boldsymbol{\beta}_c^\top \sum_{(t_k, c_k) \in \mathcal{H}_t} \boldsymbol\phi_{c_k}(t - t_k).
\label{eq:hawkes_vectorized}
\end{equation}

Let $\boldsymbol{\beta}\in\mathbb{R}^{|\mathcal C|\times|\mathcal C|}$ denote the full influence matrix whose $c$-th row equals $\boldsymbol{\beta}_c^\top$, equivalently, $\boldsymbol{\beta}
= \big[\boldsymbol{\beta}_c^\top\big]_{c=1}^{|\mathcal C|}$. To reduce quadratic parameter complexity in $\boldsymbol{\beta}$, structural assumptions such as sparsity and low rank are imposed \citep{zhou2013learning,bacry2020sparse}. Under the low-rank assumption,
\begin{equation}
\boldsymbol{\beta} = \mathbf{U}\mathbf{V}^\top,\ 
\mathbf{U}=[\mathbf{u}_c^\top\big]_{c=1}^{|\mathcal C|},\
\mathbf{V}=[\mathbf{v}_c^\top\big]_{c=1}^{|\mathcal C|},
\end{equation}
where $\mathbf{U},\mathbf{V}\in\mathbb{R}^{|\mathcal{C}|\times d}$, and $d\ll|\mathcal{C}|$. Here, $\mathbf{v}_c$ is an event embedding that captures the identity of the type of event $c$ within the sequence, a technique central to modern representation learning to map discrete elements to a continuous vector space. And $\mathbf{u}_c$ is a context embedding that describes how $c$ responds to historical influences.

Substituting the low-rank decomposition into Eq.~\eqref{eq:hawkes_vectorized} yields
\begin{align}
    \lambda_c(t | \mathcal{H}_t) &=  \mu_c + \mathbf{u}_c^\top \mathbf{V}^\top \left( \sum_{(t_k, c_k) \in \mathcal{H}_t} \boldsymbol\phi_{c_k}(t - t_k) \right)
    \nonumber\\
    &= \mu_c + \mathbf{u}_c^\top \left( \sum_{(t_k, c_k) \in \mathcal{H}_t} \mathbf{v}_{c_k} \phi_{c_k}(t - t_k) \right)\label{eq:hawkes_embedding}
\end{align}
where the inner product of the matrix $\mathbf{V}^\top$ with the sparse vector $\boldsymbol\phi_{c_k}(t - t_k)$ is equivalent to selecting the $c_k$-th column of $\mathbf{V}^\top$, which is precisely the vector of event embedding $\mathbf{v}_{c_k}$. Here, $\mathbf{v}_{c_k}$ captures the semantic representation of event type $c_k$, 
while $\phi_{c_k}(\Delta t)$ modulates its influence over time. The inner product with $\mathbf{u}_c$ measures the compatibility between the target type and the historical context. Thus, each past event is represented as a time-modulated embedding $\mathbf{v}_{c_k}\phi(\Delta t)$, and the Hawkes intensity becomes an inner product between the target context vector $\mathbf{u}_c$ and a linear history summary.

\subsection{Per-type neural influence kernels}
\label{sec:per_type_kernel}

We now address the second limitation of rigid or shared decays by introducing flexible, per-type neural influence kernels that are both more expressive and more interpretable for capturing complex, heterogeneous event dynamics. We first let each type of event $c$ have its own learnable kernel $\phi_c(\cdot)$, parameterized by a small multi-layer perceptron (MLP) applied to the elapsed time $\Delta t$. Concretely, for every event type \(c\in\mathcal C\) we define
\begin{equation}
\phi_c(\Delta t) \;=\; \mathrm{MLP}_c\big(\Delta t).
\label{eq:phi_mlp}
\end{equation}
In practice, we use a lightweight MLP with a small hidden dimension and number of layers. This guarantees both computational efficiency and interpretability of learned kernel shapes, while providing expressiveness beyond fixed parametric forms. We do not impose monotonicity or positivity on $\phi_c(\Delta t)$ because (a) the direction of influence is domain-dependent and may exhibit inhibition, and (b) learning restrictions would reduce the generality of the model, effectively reverting it to a parametric Hawkes variant.

\subsection{Hawkes Attention: time-modulated self-attention}

Eq.~\eqref{eq:hawkes_embedding} already resembles the attention pattern: the context vector $\mathbf u_c$ acts like a query while time-modulated embeddings $\mathbf v_{c_k}\phi_{c_k}(\Delta t)$ serve as a key-value pair. Temporal information is injected through decay $\phi(\Delta t)$ rather than through positional encodings. However, the linear sum cannot model context-dependent interactions among past events. Motivated by these gap, we naturally generalize by replacing the linear aggregator with a masked multi-head self-attention, which yields a nonlinear, context-aware history summary while keeping direct temporal modulation.

We let each historical event $k$ contribute a time-modulated embedding $\mathbf v_{c_k}\,\phi_{c_k}(t_j-t_k)$ to the computation of the representation at target time $t_j$. Concretely, for target index $j$ and a past index $k$ we form:
\begin{align}
Q_{j,k} &= W_Q\,\mathbf{v}_{c_j}\,\phi_{c_j}(t_j-t_k),\label{eq:Q} \\
K_{k,j} &= W_K\,\mathbf{v}_{c_k}\,\phi_{c_k}(t_j-t_k),\label{eq:K} \\
V_{k,j} &= W_V\,\mathbf{v}_{c_k}\,\phi_{c_k}(t_j-t_k).\label{eq:V}
\end{align}
where $W_Q,W_K,W_V\in\mathbb R^{d\times d}$ are learnable projections. Under the attention transformation, the theoretically derived vectors map directly to trainable attention inputs: the target context vector $\mathbf{u}_c$ naturally serves as the query, while each event semantic vector $\mathbf{v}_c$ is obtained by the standard type-to-embedding projection.

A key design choice is how time-dependent kernels influence representations. We adopt a multiplicative modulation because it directly corresponds to the intuition of the Hawkes process and our previous derivations: the elapsed time changes the strength of contribution of a past event, multiplying by $\phi_{c_k}$ modulates how influence of a source evolves with lag, and multiplying by $\phi_{c_j}$ enables the target to encode temporal receptivity. This asymmetric design is a distinctive feature of Hawkes Attention. Empirically and conceptually, multiplicative modulation yields more interpretable influence curves and aligns with classical kernel scaling.

Masked scaled dot-product attention aggregates these modulated representations to get a context representation for event $j$:
\begin{equation}
\mathbf{h}(t_j) 
= \sum_{t_k<t_j}\mathrm{softmax}_k\!\left(\frac{Q_{j,k}^\top K_{k,j}}{\sqrt{d}}\right)V_{k,j}.
\label{eq:final_h}
\end{equation}

\subsection{Multi-head architecture}
In practice, we implement multi-head attention by computing Eqs.~\eqref{eq:Q}--\eqref{eq:final_h} per head with independent projections, as in the Transformer encoder. Each head learns distinct per-type kernels $\phi_c$, enabling heterogeneous temporal behaviors.

For each head $h$, the output at time $t_j$ is given by
\begin{equation}
o_{j}^{(h)}
=
\sum_{k<j}
\operatorname{softmax}_{k}\!\left(
\frac{
\left(Q_{j,k}^{(h)}\right)^{\top} K_{k,j}^{(h)}
}{
\sqrt{d}
}
\right)
V_{k,j}^{(h)} ,
\end{equation}
where $Q_{j,k}^{(h)}$, $K_{k,j}^{(h)}$, and $V_{k,j}^{(h)}$ denote the query, key, and value representations in the $h$-th head, respectively. The final hidden representation is obtained by concatenating the outputs of all heads,
\begin{equation}
\label{eq:concateh}
\mathbf{h}(t_j)  = \operatorname{Concat}_{h=1}^{H}\!\big(o_{j}^{(h)}\big).
\end{equation}
Given the attention-produced history representation $h(t)$ from Eq.\eqref{eq:hawkes_embedding} and Eq.\eqref{eq:concateh}, the final intensity is
\begin{equation}
\label{eq:fianl_intensity}
\lambda_c(t|\mathcal{H}_t)
= \mathrm{softplus}\!\left(\mu_c + \mathbf{a}_c^\top\mathbf{h}(t)\right),
\end{equation}
where softplus ensures intensity non-negativity and stabilizes optimization, and $\mathbf{a}_c$ is a learnable vector that maps the hidden representation $\mathbf{h}(t)$ to a scalar. 

Because $\mathbf{a}_c^\top\mathbf{h}(t)$ is linear in each $o_j^{(h)}$, the contribution of a past event $k$ to $\lambda_c$ is a linear combination of its per-head kernels:
\begin{equation}
\kappa_{a\to c}^{\mathrm{eff}}(\Delta t)
=
\sum_{h=1}^H 
\bar c^{(h)}_{a\to c}\,
\phi_{a}^{(h)}(\Delta t), 
\end{equation}
where $\phi_{a}^{(h)}(\Delta t)$ is the kernel induced by the $h$-th head for events of type $a$, and
$\bar{c}_{a \to c}^{(h)}$ denotes the coefficient induced by the attention weights and the output projection.
In other words, the multi-head architecture does not learn a single kernel directly; instead, it represents a rich family of Hawkes-like excitation functions by mixing multiple head-wise kernels across event types.

\begin{figure}
\vspace{.3in}
\centerline{
\includegraphics[width=1\linewidth]{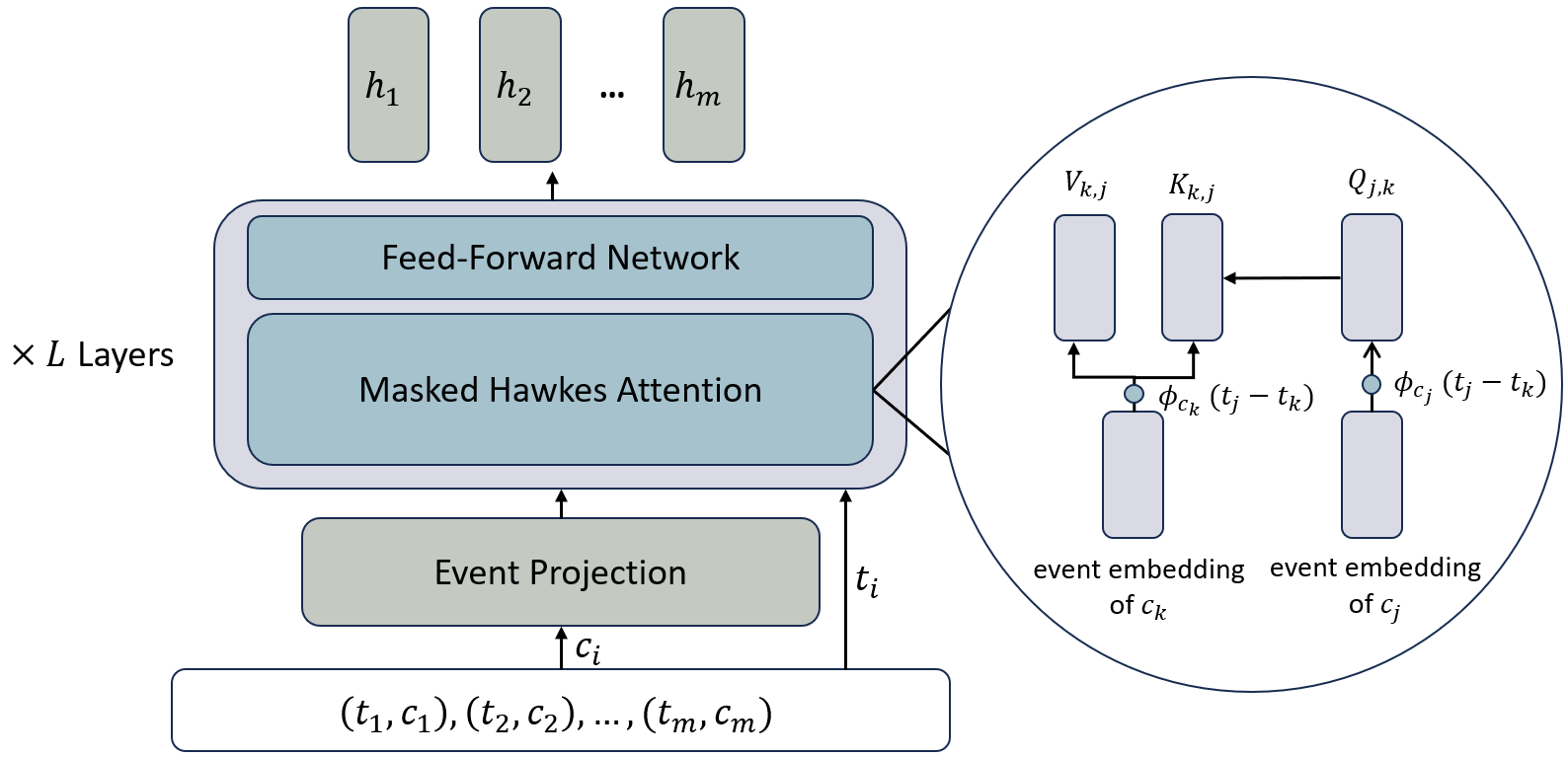}}
\caption{The complete model architecture from input sequence through event embedding, stacked Transformer-encoder layers (Feed-Forward Network after Masked Hawkes Attention) to hidden representations.}
\label{fig:archi}
\vspace{.3in}
\end{figure}

A causal mask is used to ensure that only events with $t_k<t_j$ are attended. In practice, stacking layers enables increasingly complex temporal dependencies, while an optional RNN layer can refine local dynamics in specific datasets \citep{wang2019languagemodels}. Figure \ref{fig:archi} shows an overview of our model. We intentionally adopt the simplest possible architectural blocks to isolate the contribution of the proposed time-modulated mechanism.

\subsection{Intensity and training}
The final intensity Eq.~\eqref{eq:fianl_intensity} yields an end-to-end differentiable conditional intensity model, obtained by a principled generalization of the classic Hawkes process.

The training objective is the negative log-likelihood \citep{hawkes1971spectra,mei2017neural}
\begin{equation}
\mathcal{L}
= -\sum_{i=1}^m \log \lambda_{c_i}(t_i|\mathcal{H}_{t_i})
+\int_{t_0=0}^{t_m}\sum_{c\in\mathcal C} \lambda_c(t|\mathcal{H}_t)\,dt.
\label{eq:loss}
\end{equation}
The survival integral has no closed form, so we approximate it via the Monte Carlo approximation by sampling a set of event times \citep{mei2017neural}. To predict the time and type of the next event, the minimum Bayes risk principle and the thinning algorithm are applied \citep{ogata1988thinning,xue2023easytpp}.

The model parameter memory scales as $O(|\mathcal{C}|\cdot d+|\mathcal{C}|\cdot p_{\phi}+L\cdot d^2+L\cdot d\cdot d_{\text{ff}})$, where $L$ is the number of stacked attention layers, $p_{\phi}$ is the number of parameters of each single kernel $\phi$, and $d_{\text{ff}}$ is the hidden dimension of the feed-forward networks. However, since we only use small neural networks to parameterize the kernels, $p_{\phi}$ could be regarded as constant (see Experiment \ref{exp:setup}). Therefore, our model does not incur additional parameter space complexity compared to the Transformer encoder structure. The runtime activation memory is dominated by the Transformer attention term $O(B\cdot H \cdot m^2)$ and the per-layer activations $O(B\cdot m \cdot d \cdot L)$, where $B$ is the batch size and $H$ is the number of heads. Thus, our model inherits the standard memory complexity of the Transformer encoders on sequence forecasting.

The final model unifies probabilistic interpretability with the flexibility of attention, providing a principled and scalable operator for MTPP data modeling.

\section{EXPERIMENTS}
\label{sec:exp}
In this section, we conduct a series of experiments to evaluate our proposed model in multiple real-world datasets and compare it with existing models. The experiments are designed to evaluate the effectiveness of our approach in modeling MTPP.

\subsection{Datasets}

We evaluate our model on several real-world datasets, each representing a distinct domain and temporal characteristic. These datasets contain event sequences, where each event is associated with a timestamp and a discrete type of event.
The StackOverflow dataset \citep{leskovec2014snap} contains sequences of user rewards on the StackOverflow platform, where each event records the receipt of a particular badge. The Amazon dataset \citep{ni2018amazon} comprises time-stamped product review events from January 2008 to October 2018. Each event includes a timestamp and the category of the reviewed product, and we work with a subset of 5,200 active users with \(K=16\) distinct event types. The Taxi dataset \citep{whong2014foiling} records time-stamped taxi pick-up and drop-off events across New York City’s five boroughs. Each combination (borough, pick-up / drop-off) defines an event type such that \(K=10\), and we experiment on a subset of 2,000 drivers. Finally, the Taobao dataset \citep{xue2022hypro} records the user click behavior on Taobao from November 25 to December 3, 2017. An item type categorizes each click. We retain \(K=17\) event types after preprocessing and evaluate on a subset of 4,800 active users. The statistics of each dataset are summarized in Table~\ref{tab:stats}.

\begin{table*}[t]
\caption{Statistics of Each Dataset}
\label{tab:stats}
\begin{center}
\begin{tabular}{l c r r r c c c}
\textbf{DATASET} & \textbf{$K$} & \multicolumn{3}{c}{\textbf{\# Of EVENT TOKENS}} & \multicolumn{3}{c}{\textbf{SEQUENCE LENGTH}} \\
\hline \\
&     & \textbf{Train}   & \textbf{Valid} & \textbf{Test} & \textbf{Min} & \textbf{Mean} & \textbf{Max} \\
 Taobao   & 17  & 350000  &  53000 & 101000 &   3 &  51  &  94 \\
Amazon    & 16  & 288000  &  12000 &  30000 &  14 &  44  &  94 \\
Taxi      & 10  &  51000  &   7000 &  14000 &  36 &  37  &  38 \\
StackOverflow & 22  &  90000  &  25000 &  26000 &  41 &  65  & 101 
\end{tabular}
\end{center}
\end{table*}

\subsection{Metrics}

We evaluate the models using standard metrics for MTPP modeling. We focus on the prediction of the next event: the ability of the model to predict the timestamp and the type of the next event given its history. We report Root Mean Square Error (RMSE) for time prediction and the classification error rate for type prediction. For all metrics, lower values indicate better performance.

\subsection{Baselines}

We compare our model against eight representative neural models. RMTPP and NHP follow recurrent structures such as RNN and LSTM; ODETPP utilizes ODEs for parameterization; FullyNN provides a general neural parameterization; and IFTPP bypasses explicit intensities; SAHP, THP and AttNHP are developed based on attention mechanisms. These models represent both RNN-based and attention-based methods. All hyperparameters, implementations of the benchmark model and tuning procedures are adopted from the EasyTPP suite \citep{xue2023easytpp} to ensure a consistent and fair comparison. Detailed hyperparameters are provided in the {Appendix~\ref{app:model_hyper}}.

\subsection{Experimental Setup and Training Details}
\label{exp:setup}

We perform our experiment using a uniform PyTorch \citep{paszke2019pytorch} codebase, which runs on a single NVIDIA RTX 4090 GPU. All MTPP models are implemented and evaluated within the EasyTPP framework, which provides standardized data splits, training pipelines, and faithful reimplementations of baselines. We train our model on the loss of negative log-likelihood  Eq.~\eqref{eq:loss} via the Adam optimizer \citep{kingma2015adam} with an initial learning rate chosen from \(\{10^{-3}, 5\times 10^{-4}, 10^{-4}\}\). We fix the number of attention layers \(L\in\{2,3,4\}\), number of heads \(H\in\{2,4\}\), and the hidden dimensions \(D\in\{32,64,128,256\}\). We run the experiments on five different seeds. To ensure a fair comparison, we directly report EasyTPP’s published results for all datasets except Taobao. Since our model structure is based on THP, we maintain consistency in model architecture and hyperparameters with THP, where applicable. A comparison with the THP results sufficiently demonstrates the effectiveness of our approach. We apply layer normalization \citep{ba2016layernorm}, early stopping, weight decay, and dropout as in the THP configuration. We then report the RMSE for the next event time prediction and the error rates of the event type prediction. The detailed hyperparameters of our model are provided in {Appendix~\ref{app:model_hyper}}. All other implementation details, including dataset splits, numerical integration, task setups, hyperparameter settings, etc., are available in EasyTPP. 
Note that FullyNN does not support multitype event sequences, and we did not get a reasonable result on the Taobao dataset. Therefore, it is excluded from the type prediction task and the Taobao dataset.

\subsection{Next Event Time RMSE Comparison}

We then present and analyze the performance of our model to demonstrate its effectiveness.

\begin{table*}[t]
\caption{Next Event Prediction Time RMSE Comparison}
\label{tab:rmse_comparison}
\begin{center}
\begin{tabular}{l c c c c}
\textbf{MODEL} & \multicolumn{4}{c}{\textbf{RMSE}} \\
\hline \\
& \textbf{Taxi} & \textbf{Amazon} & \textbf{StackOverflow} & \textbf{Taobao} \\
RMTPP    & 0.371 (0.003) & 0.620 (0.005) & 1.376 (0.018) & 0.133 (0.0003) \\
NHP      & 0.369 (0.003) & 0.621 (0.005) & 1.372 (0.011) & 0.147 (0.0013) \\
SAHP     & 0.372 (0.003) & 0.619 (0.005) & 1.375 (0.013) & 0.133 (0.0006) \\
AttNHP   & 0.371 (0.003) & 0.621 (0.005) & 1.372 (0.019) & 0.137 (0.0027) \\
ODETPP  & 0.371 (0.003) & 0.620 (0.006) & 1.374 (0.022) &  0.141 (0.0034) \\
FullyNN  & 0.373 (0.003) & 0.615 (0.005) & 1.375 (0.015) &  NA \\
IFTPP  & 0.373 (0.003) & 0.618 (0.005) & 1.373 (0.010) & 0.173 (0.0158) \\
THP      & 0.370 (0.003) & 0.621 (0.003) & 1.374 (0.022) & 0.134 (0.0006) \\
Ours     & \textbf{0.367} (0.002) & \textbf{0.559} (0.009) & \textbf{1.370} (0.022) & 0.134 (0.0011)
\end{tabular}
\end{center}
\end{table*}

Table~\ref{tab:rmse_comparison} compares the RMSE of the forecasts for the next event time (standard deviations in parentheses). Relative to THP, which is our direct architectural predecessor, our model produces a lower RMSE in Taxi (0.367 vs 0.370), Amazon (0.559 vs. 0.621) and StackOverflow (1.370 vs.\ 1.374), demonstrating that explicit, learnable influence kernels \(\phi_{c_k}(\Delta t)\) improve the model's ability to capture complex temporal decay patterns. Across all baselines, our RMSE is the best in three of four tasks, underscoring that modeling the temporal profile of each event type yields improved time-forecasting performance.

\subsection{Next Event Type Error Rate Comparison}

\begin{table*}[t]
\caption{Next Event Type Error Rate Comparison}
\label{tab:error_rate}
\begin{center}
\begin{tabular}{l c c c c}
\textbf{MODEL} & \multicolumn{4}{c}{\textbf{ERROR RATE}} \\
\hline \\
& \textbf{Taxi} & \textbf{Amazon} & \textbf{StackOverflow} & \textbf{Taobao} \\
RMTPP    &  9.51 (0.03) & 68.1 (0.60) & 57.3 (0.50) & 39.1 (0.13) \\
NHP      &  8.50 (0.05) & 67.1 (0.60) & 55.0 (0.60) & 44.6 (0.60) \\
SAHP     &  9.75 (0.08) & 67.7 (0.60) & 56.1 (0.50) & 56.4 (0.05) \\
AttNHP   &  8.71 (0.04) & 65.3 (0.60) & 55.2 (0.30) & 51.8 (0.97) \\
ODETPP   &  10.54 (0.08) & 65.8 (0.80) & 56.8 (0.40) & 48.0 (0.37) \\
IFTPP   &  8.56 (0.60) & 67.5 (0.70) & 55.1 (0.50) & 39.7 (0.24) \\
THP      &  8.68 (0.06) & 66.1 (0.70) & 55.0 (0.60) & 40.3 (0.21) \\
Ours     &  \textbf{8.50} (0.08) & \textbf{65.3} (0.09) & \textbf{54.6} (0.04) & \textbf{39.0} (0.15)
\end{tabular}
\end{center}
\end{table*}

Table~\ref{tab:error_rate} reports next-event type prediction error rates (standard deviations in parentheses). Against THP, our model reduces the error from 66.1\% to 65.3\% on Amazon, from 55.0\% to 54.6\% on StackOverflow, from 8.68\% to 8.50\% on Taxi and from 40.3\% to 39.0\% on Taobao, confirming that explicit time-aware kernels sharpen the semantic distinctions between event types and highlighting our model’s superior classification capacity on datasets with irregular intervals. These results demonstrate that integrating dedicated MLP kernels for each event type not only enhances temporal accuracy but also consistently improves event-type prediction across diverse real-world scenarios.

As shown in Tables~\ref{tab:stats}--\ref{tab:error_rate}, our model consistently improves the RMSE for the next event time and the error rates of the type between data sets exhibiting a wide range of characteristics, including short versus long sequence lengths and small versus large event-type vocabularies. By replacing THP’s implicit decay with explicit, per-type influence kernels \(\phi_{c_k}(\Delta t)\), our approach adapts to heterogeneous temporal dynamics without altering the core Transformer-encoder architecture. This unified, time-aware mechanism yields robust gains in forecast accuracy and classification reliability regardless of whether the underlying data are sparse point streams, dense click sequences, or long user activity logs.

\subsection{Ablation Study}

We perform an ablation study on the same datasets and tasks as before, comparing our components against variants of Hawkes Attention to demonstrate their effectiveness. We maintain the training procedures and hyperparameters unchanged.

\subsubsection{Comparing against THP}
First, our structural predecessor THP shares the same encoder-based architecture as ours, including the number of heads and layers, etc. This design follows the principle of ``equal-capacity'' comparisons so that the temporal mechanism is the only changed component. It can be viewed as a special case of our model, obtained by setting \(\phi(\Delta t)=1\) to remove multiplicative time modulation, so that queries/keys/values are not scaled by elapsed time. Instead, temporal information is injected via standard positional encodings. Consequently, THP is a natural baseline for our ablation comparisons. According to Tables~\ref{tab:rmse_comparison} and Table~\ref{tab:error_rate}, our model attains better performance, suggesting that learning per-type multiplicative influence kernels can capture richer, type-specific temporal dynamics and information than standard positional encodings. 

\subsubsection{Per-type kernel}

\begin{table*}[h]
\begin{center}
\caption{Ablation Study on Per-type Kernels}
\label{tab:ablation_kernel}
\begin{tabular}{l c c c c}
\textbf{MODEL} & \multicolumn{4}{c}{\textbf{RMSE / ERROR RATE (\%)}} \\
\hline \\
& \textbf{Taxi} & \textbf{Amazon} & \textbf{StackOverflow} & \textbf{Taobao} \\
\multirow{2}{*}{Original} 
& 0.367/\textbf{8.50} & 0.559/65.3  & \textbf{1.370/54.6} & \textbf{0.134/39.0}\\
& (0.002/0.08) & (0.009/0.09) & (0.022/0.04) & (0.0011/0.15)\\
\multirow{2}{*}{Shared $\phi$} 
& 0.367/8.60 & 0.553/65.3 & 1.410/54.8 & 0.137/39.1  \\
& (0.003/0.08) & (0.014/0.23) & (0.038/0.29) & (0.0011/0.08)
\end{tabular}
\end{center}
\end{table*}
We then perform ablation on the Hawkes Attention variant by replacing the per-type MLP kernels with a single shared kernel $\phi$ for all types of events. From Table~\ref{tab:ablation_kernel} we observe that the per-type kernels yield improvements on several metrics: notably lower RMSE on StackOverflow (1.370 vs.\ 1.410) and Taobao (0.134 vs.\ 0.137), and slightly better type error rates on Taxi and Taobao. The shared kernel variant achieves a marginally better RMSE on Amazon (0.553 vs.\ 0.559) but with larger run-to-run variance, suggesting the shared kernel can sometimes generalize better on particular data regimes, such as when per-type data are sparse, at the cost of losing type-specific temporal nuance. Overall, these results indicate that per-type temporal kernels tend to improve temporal and type prediction accuracy in practice, especially on datasets where different event types exhibit distinct temporal profiles, while the shared kernel offers a cheaper but less expressive alternative.

\subsubsection{Positional encodings}

\begin{table*}[t]
\begin{center}
\caption{Ablation Study on Positional Encodings (PE)}
\label{tab:ablation_pe}
\begin{tabular}{l c c c c}
\textbf{MODEL} & \multicolumn{4}{c}{\textbf{RMSE / ERROR RATE (\%)}} \\
\hline \\
& \textbf{Taxi} & \textbf{Amazon} & \textbf{StackOverflow} & \textbf{Taobao} \\
\multirow{2}{*}{Original} 
& 0.367/\textbf{8.50} & 0.559/65.3 & \textbf{1.370}/54.6 & \textbf{0.134/39.0}\\
& (0.002/0.08) & (0.009/0.09) & (0.022/0.04) & (0.0011/0.15)\\
\multirow{2}{*}{Original+PE} 
& 0.365/8.59 & 0.555/65.3 & 1.382/54.5 & 0.136/39.1 \\
& (0.002/0.07) & (0.009/0.08) & (0.018/0.06) & (0.0009/0.14)
\end{tabular}
\end{center}
\end{table*}

We further ablate the classical Transformer positional encodings (see {Appendix~\ref{app:attention}}) by adding them to our full Hawkes-Attention model. The results are shown in Table~\ref{tab:ablation_pe}. Adding standard positional encodings does not yield consistent or meaningful improvement (changes are small and fall within run-to-run variance), indicating that our learned kernels \(\phi_c(\Delta t)\) already capture the relevant temporal information that positional encodings would provide. In other words, \(\phi_c(\Delta t)\) effectively substitutes for classical positional encodings and offers more flexibility and interpretability. 

\subsection{Interpretability of the Influence Kernels}

\begin{figure}[t]
    \centering
    \includegraphics[width=0.9\linewidth]{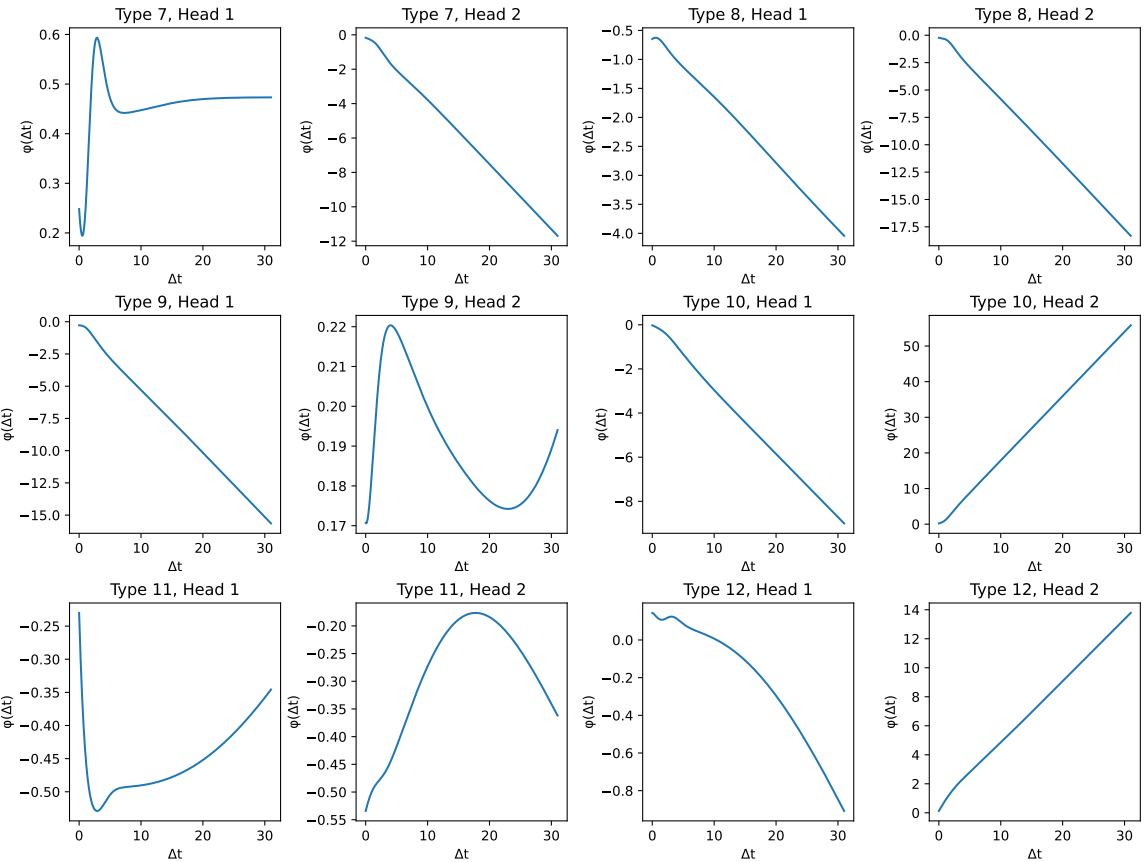}
    \caption{Sample $\phi$ curves of different event types}
    \label{fig:phi_curve}
\end{figure}

Taking StackOverflow as a case study, we illustrate how the learned influence kernels \(\phi_{c}(\Delta t)\) accept intuitive interpretations. Each type of badge \(c\) corresponds to a different \(\phi_{c}\), and Figure \ref{fig:phi_curve} shows some learned kernel curves $\phi_c(\Delta t)$, demonstrating different temporal dynamics of user engagement.

Many badges exhibit a monotonically decaying \(\phi(\Delta t)\), indicating a strong, short-lived boost in activity immediately after award that gradually fades. Other kernels plateau after an initial drop, suggesting a sustained moderate effect once the user overcomes an initial novelty period. In several instances, the \(\phi\) curves show a brief local minimum followed by a rebound, capturing a temporary slowdown before reengaging more actively, perhaps due to the cooling effect. In contrast, some badges produce an increasing \(\phi\) over time, reflecting rewards that encourage accumulating behavior or longer-term preparation before the next contribution.

The two attention heads decompose each badge’s impact into complementary components: one head typically learns a smooth and long-term trend (linear or sublinear growth or decay), while the other captures transient fluctuations (peaks, dips, or convexity) on shorter time scales. Together, they enable the model to represent both the enduring and the ephemeral influences of each badge.

In general, these patterns in \(\phi_{c}(\Delta t)\) align with the real-world role of badges on StackOverflow, offering interpretable insights into how different rewards stimulate user behavior over time.

\section{DISCUSSION AND CONCLUSION}
\label{sec:dis_conclu}
In this section, we first extend our model originally designed for MTPP to a broader sequence modeling framework. By interpreting time series through the lens of the Hawkes process, we demonstrate that our attention mechanism can handle both evenly spaced and irregularly spaced timestamps, as well as discrete and continuous values. This extension bridges the gap between MTPP and traditional time series forecasting, forming a unified framework for sequential data modeling.

Our approach generalizes to time series by treating regular measurements as a special case of the same framework of asynchronous timestamps. This allows our model to adapt to different types of temporal data naturally, providing a consistent and mathematically grounded method for handling diverse sequential data tasks. The ability to seamlessly integrate discrete or continuous variables and to model both equally spaced and irregularly spaced timestamps positions our model as a versatile tool for a wide range of sequence-based tasks.

We further demonstrate the strength of this general framework by applying it to traditional time series forecasting problems, achieving competitive results across a variety of datasets. This shows that our method can be effectively generalized beyond MTPP, offering a robust solution for sequential data modeling in both MTPP and time series domains. Details of the model and experiments can be found in {Appendix~\ref{app:time-series}}.

Finally, in this paper, we presented a novel approach for modeling MTPP using an attention mechanism mathematically derived from the Hawkes process. Unlike traditional methods, which rely on predefined influence kernels, our model uses neural networks to model event-specific influence kernels, providing greater flexibility and expressiveness. Additionally, by using time differences as input and leveraging event embeddings, our attention mechanism remains time-aware even without the need for positional encodings, making it suitable for a wide range of asynchronous sequential data tasks.

We demonstrated the effectiveness of our approach through experiments with MTPP tasks. Our model consistently outperforms existing MTPP methods in different metrics, showing its strong performance in real-world applications. 

\section*{Acknowledgments}

The authors would like to thank the anonymous reviewers for their helpful comments.

This work was supported by the MOE AcRF Tier 1 Grant A-8003569-00-00 and the NUS Start-up Grant A-0009985-00-00.

\begingroup
\raggedright
\sloppy
\emergencystretch=2em
\bibliographystyle{chicago}
\bibliography{refs}

@inproceedings{du2016recurrent,
  author       = {Du, Nan and Dai, Hanjun and Trivedi, Ritwik Kumar and Upadhyay, Utkarsh and Gomez-Rodriguez, Manuel and Song, Le},
  title        = {Recurrent Marked Temporal Point Processes: Embedding Event History to Vector},
  booktitle    = {Proceedings of the ACM SIGKDD International Conference on Knowledge Discovery and Data Mining},
  year         = {2016},
  publisher    = {ACM},
}

@inproceedings{kingma2015adam,
  author       = {Kingma, Diederik P. and Ba, Jimmy Lei},
  title        = {Adam: A Method for Stochastic Optimization},
  booktitle    = {International Conference on Learning Representations},
  year         = {2015},
}

@misc{leskovec2014snap,
  author       = {Leskovec, Jure and Krevl, Andrej},
  title        = {{SNAP} Datasets: Stanford Large Network Dataset Collection},
  year         = {2014},
  howpublished = {\newline\url{http://snap.stanford.edu/data}},
}

@phdthesis{liniger2009multivariate,
  title={Multivariate hawkes processes},
  author={Liniger, Thomas},
  year={2009},
  school={ETH Zurich}
}

@article{mei2017neural,
  title={The neural hawkes process: A neurally self-modulating multivariate point process},
  author={Mei, Hongyuan and Eisner, Jason M},
  journal={Advances in neural information processing systems},
  volume={30},
  year={2017}
}

@misc{ni2018amazon,
  author       = {Ni, Jianmo},
  title        = {Amazon Review Data},
  year         = {2018},
  howpublished = {\newline\url{https://nijianmo.github.io/amazon/}},
}

@article{paszke2019pytorch,
  title={Pytorch: An imperative style, high-performance deep learning library},
  author={Paszke, Adam and Gross, Sam and Massa, Francisco and Lerer, Adam and Bradbury, James and Chanan, Gregory and Killeen, Trevor and Lin, Zeming and Gimelshein, Natalia and Antiga, Luca and others},
  journal={Advances in neural information processing systems},
  volume={32},
  year={2019}
}

@misc{whong2014foiling,
  author       = {Whong, C.},
  title        = {FOILing NYC’s Taxi Trip Data},
  year         = {2014},
  howpublished = {\newline\url{https://github.com/whong/nyc-taxi-data}},
}

@inproceedings{yang2022transformer,
  title={Transformer Embeddings of Irregularly Spaced Events and Their Participants},
  author={Yang, Chenghao and Mei, Hongyuan and Eisner, Jason},
  booktitle={Proceedings of the Tenth International Conference on Learning Representations},
  year={2022}
}

@inproceedings{zhang2020self,
  title={Self-attentive Hawkes process},
  author={Zhang, Qiang and Lipani, Aldo and Kirnap, Omer and Yilmaz, Emine},
  booktitle={International conference on machine learning},
  pages={11183--11193},
  year={2020},
  organization={PMLR}
}

@inproceedings{zuo2020transformer,
  title={Transformer hawkes process},
  author={Zuo, Simiao and Jiang, Haoming and Li, Zichong and Zhao, Tuo and Zha, Hongyuan},
  booktitle={International conference on machine learning},
  pages={11692--11702},
  year={2020},
  organization={PMLR}
}

@article{xue2022hypro,
  title={Hypro: A hybridly normalized probabilistic model for long-horizon prediction of event sequences},
  author={Xue, Siqiao and Shi, Xiaoming and Zhang, James and Mei, Hongyuan},
  journal={Advances in Neural Information Processing Systems},
  volume={35},
  pages={34641--34650},
  year={2022}
}

@article{ba2016layernorm,
  title={Layer Normalization},
  author={Ba, Jimmy Lei and Kiros, Jamie Ryan and Hinton, Geoffrey E},
  journal={stat},
  volume={1050},
  pages={21},
  year={2016}
}

@article{wu2021autoformer,
  title={Autoformer: Decomposition transformers with auto-correlation for long-term series forecasting},
  author={Wu, Haixu and Xu, Jiehui and Wang, Jianmin and Long, Mingsheng},
  journal={Advances in neural information processing systems},
  volume={34},
  pages={22419--22430},
  year={2021}
}

@article{liu2022scinet,
  title={Scinet: Time series modeling and forecasting with sample convolution and interaction},
  author={Liu, Minhao and Zeng, Ailing and Chen, Muxi and Xu, Zhijian and Lai, Qiuxia and Ma, Lingna and Xu, Qiang},
  journal={Advances in Neural Information Processing Systems},
  volume={35},
  pages={5816--5828},
  year={2022}
}

@inproceedings{zhou2022fedformer,
  title={Fedformer: Frequency enhanced decomposed transformer for long-term series forecasting},
  author={Zhou, Tian and Ma, Ziqing and Wen, Qingsong and Wang, Xue and Sun, Liang and Jin, Rong},
  booktitle={International conference on machine learning},
  pages={27268--27286},
  year={2022},
  organization={PMLR}
}

@article{liu2022nonstationary,
  title={Non-stationary transformers: Exploring the stationarity in time series forecasting},
  author={Liu, Yong and Wu, Haixu and Wang, Jianmin and Long, Mingsheng},
  journal={Advances in neural information processing systems},
  volume={35},
  pages={9881--9893},
  year={2022}
}

@inproceedings{
zhang2023crossformer,
title={Crossformer: Transformer Utilizing Cross-Dimension Dependency for Multivariate Time Series Forecasting},
author={Yunhao Zhang and Junchi Yan},
booktitle={The Eleventh International Conference on Learning Representations },
year={2023},
url={https://openreview.net/forum?id=vSVLM2j9eie}
}

@inproceedings{
nie2023time,
title={A Time Series is Worth 64 Words:  Long-term Forecasting with Transformers},
author={Yuqi Nie and Nam H Nguyen and Phanwadee Sinthong and Jayant Kalagnanam},
booktitle={The Eleventh International Conference on Learning Representations },
year={2023},
url={https://openreview.net/forum?id=Jbdc0vTOcol}
}

@inproceedings{zeng2023transformers,
  title={Are transformers effective for time series forecasting?},
  author={Zeng, Ailing and Chen, Muxi and Zhang, Lei and Xu, Qiang},
  booktitle={Proceedings of the AAAI conference on artificial intelligence},
  volume={37},
  number={9},
  pages={11121--11128},
  year={2023}
}

@article{
das2023tide,
title={Long-term Forecasting with Ti{DE}: Time-series Dense Encoder},
author={Abhimanyu Das and Weihao Kong and Andrew Leach and Shaan K Mathur and Rajat Sen and Rose Yu},
journal={Transactions on Machine Learning Research},
issn={2835-8856},
year={2023},
url={https://openreview.net/forum?id=pCbC3aQB5W},
note={}
}

@article{li2023revisiting,
  title={Revisiting long-term time series forecasting: An investigation on linear mapping},
  author={Li, Zhe and Qi, Shiyi and Li, Yiduo and Xu, Zenglin},
  journal={arXiv preprint arXiv:2305.10721},
  year={2023}
}

@inproceedings{
wu2023timesnet,
title={TimesNet: Temporal 2D-Variation Modeling for General Time Series Analysis},
author={Haixu Wu and Tengge Hu and Yong Liu and Hang Zhou and Jianmin Wang and Mingsheng Long},
booktitle={The Eleventh International Conference on Learning Representations },
year={2023},
url={https://openreview.net/forum?id=ju_Uqw384Oq}
}

@book{daley2008introduction,
  author       = {Daley, Daryl J. and Vere‐Jones, David},
  title        = {An Introduction to the Theory of Point Processes, Volume II:General Theory and Structure},
  publisher    = {Springer},
  year         = {2008},
}

@article{hawkes1971spectra,
  author       = {Hawkes, Alan G.},
  title        = {Spectra of Some Self‐Exciting and Mutually Exciting Point Processes},
  journal      = {Biometrika},
  volume       = {58},
  number       = {1},
  pages        = {83--90},
  year         = {1971},
  publisher    = {Oxford University Press},
}

@inproceedings{wang2018supervised,
  title={Supervised reinforcement learning with recurrent neural network for dynamic treatment recommendation},
  author={Wang, Lu and Zhang, Wei and He, Xiaofeng and Zha, Hongyuan},
  booktitle={Proceedings of the 24th ACM SIGKDD international conference on knowledge discovery \& data mining},
  pages={2447--2456},
  year={2018}
}

@inproceedings{yang2011like,
  title={Like like alike: joint friendship and interest propagation in social networks},
  author={Yang, Shuang-Hong and Long, Bo and Smola, Alex and Sadagopan, Narayanan and Zheng, Zhaohui and Zha, Hongyuan},
  booktitle={Proceedings of the 20th international conference on World wide web},
  pages={537--546},
  year={2011}
}

@article{bacry2015hawkes,
  title={Hawkes processes in finance},
  author={Bacry, Emmanuel and Mastromatteo, Iacopo and Muzy, Jean-Fran{\c{c}}ois},
  journal={Market Microstructure and Liquidity},
  volume={1},
  number={01},
  pages={1550005},
  year={2015},
  publisher={World Scientific}
}

@article{irani2024positional,
  title={Positional Encoding in Transformer-Based Time Series Models: A Survey},
  author={Irani, Habib and Metsis, Vangelis},
  journal={dynamics},
  volume={4},
  pages={24},
  year={2025}
}

@article{wang2019languagemodels,
  title={Language models with transformers},
  author={Wang, Chenguang and Li, Mu and Smola, Alexander J},
  journal={arXiv preprint arXiv:1904.09408},
  year={2019}
}

@inproceedings{
xue2023easytpp,
title={Easy{TPP}: Towards Open Benchmarking Temporal Point Processes},
author={Siqiao Xue and Xiaoming Shi and Zhixuan Chu and Yan Wang and Hongyan Hao and Fan Zhou and Caigao JIANG and Chen Pan and James Y. Zhang and Qingsong Wen and JUN ZHOU and Hongyuan Mei},
booktitle={The Twelfth International Conference on Learning Representations},
year={2024},
url={https://openreview.net/forum?id=PJwAkg0z7h}
}

@inproceedings{
liu2023itransformer,
title={iTransformer: Inverted Transformers Are Effective for Time Series Forecasting},
author={Yong Liu and Tengge Hu and Haoran Zhang and Haixu Wu and Shiyu Wang and Lintao Ma and Mingsheng Long},
booktitle={The Twelfth International Conference on Learning Representations},
year={2024},
url={https://openreview.net/forum?id=JePfAI8fah}
}

@inproceedings{li2021informer,
  title={Informer: Beyond efficient transformer for long sequence time-series forecasting},
  author={Zhou, Haoyi and Zhang, Shanghang and Peng, Jieqi and Zhang, Shuai and Li, Jianxin and Xiong, Hui and Zhang, Wancai},
  booktitle={Proceedings of the AAAI conference on artificial intelligence},
  volume={35},
  number={12},
  pages={11106--11115},
  year={2021}
}

@inproceedings{pang2021cehr,
  title={CEHR-BERT: Incorporating temporal information from structured EHR data to improve prediction tasks},
  author={Pang, Chao and Jiang, Xinzhuo and Kalluri, Krishna S and Spotnitz, Matthew and Chen, RuiJun and Perotte, Adler and Natarajan, Karthik},
  booktitle={Machine Learning for Health},
  pages={239--260},
  year={2021},
  organization={PMLR}
}

@inproceedings{
chen2021neural,
title={Neural Spatio-Temporal Point Processes},
author={Ricky T. Q. Chen and Brandon Amos and Maximilian Nickel},
booktitle={International Conference on Learning Representations},
year={2021},
url={https://openreview.net/forum?id=XQQA6-So14}
}

@article{omi2019fullynn,
  title={Fully neural network based model for general temporal point processes},
  author={Omi, Takahiro and Aihara, Kazuyuki and others},
  journal={Advances in neural information processing systems},
  volume={32},
  year={2019}
}

@inproceedings{
shchur2020intensityfree,
title={Intensity-Free Learning of Temporal Point Processes},
author={Oleksandr Shchur and Marin Biloš and Stephan Günnemann},
booktitle={International Conference on Learning Representations},
year={2020},
url={https://openreview.net/forum?id=HygOjhEYDH}
}

@article{bacry2020sparse,
  title={Sparse and low-rank multivariate Hawkes processes},
  author={Bacry, Emmanuel and Bompaire, Martin and Ga{\"\i}ffas, St{\'e}phane and Muzy, Jean-Francois},
  journal={Journal of Machine Learning Research},
  volume={21},
  number={50},
  pages={1--32},
  year={2020}
}

@inproceedings{zhang2024neural,
  title={Neural jump-diffusion temporal point processes},
  author={Zhang, Shuai and Zhou, Chuan and Liu, Yang Aron and Zhang, Peng and Lin, Xixun and Ma, Zhi-Ming},
  booktitle={Forty-first International Conference on Machine Learning},
  year={2024}
}

@article{NEURIPS2023_4eb2c0ad,
  title={Integration-free training for spatio-temporal multimodal covariate deep kernel point processes},
  author={Zhang, Yixuan and Kong, Quyu and Zhou, Feng},
  journal={Advances in Neural Information Processing Systems},
  volume={36},
  pages={25031--25049},
  year={2023}
}

@article{NEURIPS2023_3c129892,
  title={Prompt-augmented temporal point process for streaming event sequence},
  author={Xue, Siqiao and Wang, Yan and Chu, Zhixuan and Shi, Xiaoming and Jiang, Caigao and Hao, Hongyan and Jiang, Gangwei and Feng, Xiaoyun and Zhang, James and Zhou, Jun},
  journal={Advances in Neural Information Processing Systems},
  volume={36},
  pages={18885--18905},
  year={2023}
}

@inproceedings{
bae2023meta,
title={Meta Temporal Point Processes},
author={Wonho Bae and Mohamed Osama Ahmed and Frederick Tung and Gabriel L. Oliveira},
booktitle={The Eleventh International Conference on Learning Representations },
year={2023},
url={https://openreview.net/forum?id=QZfdDpTX1uM}
}

@inproceedings{zhou2013learning,
  title={Learning social infectivity in sparse low-rank networks using multi-dimensional hawkes processes},
  author={Zhou, Ke and Zha, Hongyuan and Song, Le},
  booktitle={Artificial intelligence and statistics},
  pages={641--649},
  year={2013},
  organization={PMLR}
}

@article{ogata1988thinning,
  title={Statistical models for earthquake occurrences and residual analysis for point processes},
  author={Ogata, Yosihiko},
  journal={Journal of the American Statistical association},
  volume={83},
  number={401},
  pages={9--27},
  year={1988},
  publisher={Taylor \& Francis}
}
\endgroup

\section*{Checklist}

\begin{enumerate}

  \item For all models and algorithms presented, check if you include:
  \begin{enumerate}
    \item A clear description of the mathematical setting, assumptions, algorithm, and/or model. [Yes]
    \item An analysis of the properties and complexity (time, space, sample size) of any algorithm. [Yes]
    \item (Optional) Anonymized source code, with specification of all dependencies, including external libraries. [Yes]
  \end{enumerate}

  \item For any theoretical claim, check if you include:
  \begin{enumerate}
    \item Statements of the full set of assumptions of all theoretical results. [Not Applicable]
    \item Complete proofs of all theoretical results. [Not Applicable]
    \item Clear explanations of any assumptions. [Not Applicable]
  \end{enumerate}

  \item For all figures and tables that present empirical results, check if you include:
  \begin{enumerate}
    \item The code, data, and instructions needed to reproduce the main experimental results (either in the supplemental material or as a URL). [Yes]
    \item All the training details (e.g., data splits, hyperparameters, how they were chosen). [Yes]
    \item A clear definition of the specific measure or statistics and error bars (e.g., with respect to the random seed after running experiments multiple times). [Yes]
    \item A description of the computing infrastructure used. (e.g., type of GPUs, internal cluster, or cloud provider). [Yes]
  \end{enumerate}

  \item If you are using existing assets (e.g., code, data, models) or curating/releasing new assets, check if you include:
  \begin{enumerate}
    \item Citations of the creator If your work uses existing assets. [Yes]
    \item The license information of the assets, if applicable. [Yes]
    \item New assets either in the supplemental material or as a URL, if applicable. [Yes]
    \item Information about consent from data providers/curators. [Yes]
    \item Discussion of sensible content if applicable, e.g., personally identifiable information or offensive content. [Not Applicable]
  \end{enumerate}

  \item If you used crowdsourcing or conducted research with human subjects, check if you include:
  \begin{enumerate}
    \item The full text of instructions given to participants and screenshots. [Not Applicable]
    \item Descriptions of potential participant risks, with links to Institutional Review Board (IRB) approvals if applicable. [Not Applicable]
    \item The estimated hourly wage paid to participants and the total amount spent on participant compensation. [Not Applicable]
  \end{enumerate}

\end{enumerate}

\clearpage
\appendix
\thispagestyle{empty}

\onecolumn
\aistatstitle{From Hawkes Processes to Attention: Time-Modulated Mechanisms for Event Sequences: \\
Supplementary Materials}

\section{ATTENTION MECHANISM AND POSITIONAL ENCODING}
\label{app:attention}

The self-attention mechanism used in Transformers has been proven to be highly effective in capturing long-range dependencies in sequential data. In standard attention mechanisms, queries, keys, and values are projected from input representations, and attention weights are computed based on their pairwise similarity. The attention output for a query \( q \) is computed as a weighted sum of the values \( v \), with the weights derived from the compatibility between \( q \) and the keys \( k \) using a similarity measure. Mathematically, the attention mechanism can be described as:
\begin{align}
    \text{Attention}(Q, K, V) = \text{softmax}\left(\frac{QK^\top}{\sqrt{d_k}}\right) V
\end{align}
where \( Q \) is the matrix of queries, \( K \) is the matrix of keys, \( V \) is the matrix of values and \( d_k \) is the dimension of the keys. 

Positional encodings are introduced to inject information about the position of tokens in a sequence for the attention mechanism. The positional encoding used in Transformers is defined as a vector where the \(i\)-th dimension is given by the following formulas:
\begin{align}
    \text{Positional Encoding}_i = 
    \begin{cases}
    \sin\left(\frac{i}{10000^{2j/d}}\right) & \text{if } j \text{ is even} \\
    \cos\left(\frac{i}{10000^{2j/d}}\right) & \text{if } j \text{ is odd}
    \end{cases}
\end{align}
where \(i\) is the position of the token, \(j\) is the dimension, and \(d\) is the total dimensionality of the embedding.

These encodings are typically added to the input embeddings as follows:
\begin{align}
    \text{Input}_i = \text{Embedding}_i + \text{Positional Encoding}_i
\end{align}
where \( \text{Embedding}_i \) is the embedding of the \(i\)-th token, and \( \text{Positional Encoding}_i \) provides a unique encoding that reflects the position of the token in the sequence. This ensures that the model remains sensitive to the order and timing of events. However, this positional encoding was originally developed for textual data and conveys only discrete token positions, thereby failing to capture the rich temporal information present in MTPP data.

\section{MODEL HYPERPARAMETERS}
\label{app:model_hyper}

\begin{table}[h]
\begin{center}
\caption{Hyperparameters of Baselines}
\label{tab:hyperparams}
\begin{tabular}{@{} l c c c c c c c c @{}}
\textbf{PARAMETER} & \textbf{RMTPP} & \textbf{NHP} & \textbf{SAHP} & \textbf{THP} & \textbf{AttNHP} & \textbf{ODETPP} & \textbf{FullyNN} & \textbf{IFTPP}\\
\hline\ \
$\texttt{d\_model}$    & 32 & 64 & 32 & 64 & 32 & 32 & 32 & 32\\
$\texttt{time\_emb\_size}$ & 16 & 16 & 16 & 16 & 16 & 16 & 16 & 16\\
$\texttt{num\_layers}$     & 2  & 2  & 2  & 2  & 1 & 2 & 2 & 2 \\
$\texttt{num\_heads}$      & NA & NA & 2 & 2 & 2 & NA & NA & NA
\end{tabular}
\end{center}
\end{table}

\begin{table}[h]
\begin{center}
\caption{Hyperparameters of Our Model Structure}
\begin{tabular}{l c @{}}
\textbf{PARAMETER} & \textbf{VALUE} \\
\hline \\
$\texttt{d\_inner\ (hidden\_size)}$   & 128 \\
$\texttt{d\_model}$                   & 64 \\
$\texttt{d\_k, d\_v}$                 & 64 \\
$\texttt{num\_layers}$                & 2 \\
$\texttt{num\_heads}$                 & 2 
\end{tabular}
\end{center}
\end{table}

\begin{table}[h]
\begin{center}         
\caption{Hyperparameters of Our Model on MTPP Datasets}
\begin{tabular}{ l *{4}{c} }
\textbf{PARAMETER} & \textbf{TAXI} & \textbf{AMAZON} & \textbf{STACKOVERFLOW} & \textbf{TAOBAO}\\
\hline \\
$\texttt{decay}$  & $1 \times 10^{-3}$  & $1 \times 10^{-3}$  & $1 \times 10^{-2}$  & $1 \times 10^{-3}$ \\
$\texttt{d\_rnn}$   & NA & 64 & 256 & 64\\
$\texttt{phi\_width}$ & 8 & 8 & 4 & 8 \\
$\texttt{phi\_depth}$   & 2 & 2 & 2 & 2 \\
$\texttt{batch\_size}$   & 256 & 256 & 256 & 256  \\
$\texttt{learning\_rate}$  & $1 \times 10^{-4}$  & $1 \times 10^{-3}$  & $1 \times 10^{-3}$  & $1 \times 10^{-3}$ \\
\end{tabular}
\end{center}
\end{table}

\section{EXTENSION ON TIME SERIES}
\label{app:time-series}
\subsection{Related Work}
\paragraph{Attention-based Time Series Forecasting}
A rich literature on forecasting regularly sampled signals has produced many high-performing architectures. Transformer variants specifically tailored to time series include Autoformer \citep{wu2021autoformer}, which decomposes trend and seasonality; FEDformer \citep{zhou2022fedformer}, which leverages frequency-domain attention for long horizons; Stationary Transformer variants that aim to utilize stationarity \citep{liu2022nonstationary}; Crossformer \citep{zhang2023crossformer} introduces cross-time and cross-dimension attention to better capture multivariate dependencies; PatchTST \citep{nie2023time} treats subsequences as patches and applies channel-independent attention to efficiently model long-range temporal patterns; and the inverted or dimension-wise design of iTransformer \citep{liu2023itransformer}. These attention-based methods excel at capturing dependencies in regularly sampled data, and often achieve state-of-the-art performance. However, they typically rely on assumptions, including regular sampling, positional encodings, and stationarity, that make them less natural for asynchronous event streams or heterogeneous mark spaces.

\paragraph{Other Time Series Models}
Complementary paradigms include linear and decomposition methods such as DLinear \citep{zeng2023transformers}, TiDE \citep{das2023tide}, RLinear \citep{li2023revisiting}, which show that carefully engineered linear components can be highly competitive and efficient, and convolutional models like SCINet \citep{liu2022scinet} and TimesNet \citep{wu2023timesnet} that focus on local and multiscale motifs. These approaches trade off expressivity, inductive bias, and efficiency in different ways: linear models are efficient but may miss complex nonlinear interactions; convolutional models capture local structure well but can struggle with event-driven dependencies. Thus, while powerful on time series tasks, these approaches do not naturally provide a unified treatment of asynchronous, marked event streams and continuous signals.

\subsection{Time Series Variant Model}
\subsubsection{Generalization from MTPP Model}
Our approach can be naturally extended to time series, which, like marked temporal point processes (MTPP), are sequence modeling problems rich in temporal information. The key distinction between time series and MTPP lies in the type of variables: while MTPP deals with discrete event types (marks), time series models typically involve continuous variables. Despite this difference, we can apply the same attention mechanism to time series by adapting it to handle continuous-valued sequences. Specifically, the fundamental structure of the attention mechanism remains unchanged, but instead of modeling influence kernels for each discrete event type, we use separate MLPs for the query, key, and value spaces to model the influence kernels for the continuous-valued time series data.

In this case, the attention formula becomes:
\begin{equation}
    Q_{j,k} = W_Q \mathbf{v}_{j}  \phi_{Q}(t_j - t_k), \quad K_{k,j} = W_K \mathbf{v}_{k}  \phi_{K}(t_j - t_k), \quad V_{k,j} = W_V \mathbf{v}_{k}  \phi_{V}(t_j - t_k)
    \label{eq:timeseries_extension}
\end{equation}
We apply separate MLPs to the query, key, and value components, effectively learning different influence kernels for each of these spaces. As before, the temporal information is captured by the difference \(\Delta t = t_j - t_k\), ensuring that the model remains time-aware even without relying on positional encodings. This allows us to extend the attention mechanism from the MTPP to continuous-valued time series while preserving its ability to model complex temporal dependencies. Because we use \(\Delta t\) and inherently fuse event embeddings, our attention mechanism is capable of handling sequences with evenly or irregularly spaced, discrete or continuous variates, thereby forming a unified and general framework for sequential data modeling.

\subsubsection{Model on Time Series}
As mentioned above, our framework extends directly to time series data, whether sampled evenly or irregularly, with only minor modifications in Eq.~\eqref{eq:timeseries_extension}. Instead of discrete event embeddings, we feed raw observations at each timestamp into the same multi-head self-attention layers, using the elapsed time differences to modulate queries, keys, and values via shared temporal kernel networks. We adopt an encoder-only architecture and do not include a decoder, since our task focuses on generating contextualized representations of past measurements rather than autoregressive generation. The encoder outputs (i.e. sequence hidden representation) are then mapped to the target time series dimension and length through a simple linear projection, producing a unified model that handles both asynchronous event streams and continuous signals without positional encodings.

\subsection{Experiment on Time Series Datasets}
\subsubsection{Time Series Datasets}
 
We evaluate our time series variant on eight widely used multivariate benchmarks. The Weather dataset \citep{wu2021autoformer} contains 21 meteorological indicators (e.g. humidity and temperature) measured hourly across locations in Germany. The Traffic dataset \citep{wu2021autoformer} records the occupancy rates of the roads of multiple loop sensors on San Francisco highways at 10-minute intervals. The ECL (Electricity) dataset \citep{wu2021autoformer} tracks the hourly power consumption of 321 customers. The Exchange-rate dataset \citep{wu2021autoformer} logs daily exchange rates for eight currencies from 1990 to 2016. Finally, the ETT (Electricity Transformer Temperature) collection \citep{li2021informer} comprises four series: two transformers each sampled at 15-minute (\textit{m}) and 1-hour (\textit{h}) resolutions, denoted \textit{ETTm1}, \textit{ETTm2}, \textit{ETTh1}, and \textit{ETTh2}. 

\subsubsection{Metrics}
For the time series experiments, we report point forecasting metrics, namely Mean Squared Error (MSE) and Mean Absolute Error (MAE), which quantify the average squared and absolute differences between predicted and true values. Lower values indicate better performance.

\subsubsection{Time Series Baselines}

To evaluate our model on time series forecasting, we compare it against eleven state-of-the-art or widely used benchmarks that span several architectural paradigms. Among Transformer-style methods we include Autoformer \citep{wu2021autoformer}, which adopts a decomposition architecture to separate trend and seasonality; FEDformer \citep{zhou2022fedformer}, which leverages frequency-domain attention to capture long-term dependencies; the Stationary Transformer \citep{liu2022nonstationary}, designed to improve stability by enforcing stationarity; Crossformer \citep{zhang2023crossformer}, which exploits cross-dimensional attention patterns; PatchTST \citep{nie2023time}, which adapts patching strategies to time series; and iTransformer \citep{liu2023itransformer}, which applies dimension-wise or “inverted” attention. We also include three recent linear-based approaches, DLinear \citep{zeng2023transformers}, a decomposition-based linear model; TiDE \citep{das2023tide}, which uses time-delay embedding for linear forecasting; and RLinear \citep{li2023revisiting}, which refines simple linear maps via residual learning. Finally, we include two convolutional methods, SCINet \citep{liu2022scinet}, which uses interactive downsampling and up-sampling convolutional blocks, and TimesNet \citep{wu2023timesnet}, which reshapes series in 2D and applies joint temporal and feature-wise convolutions. By benchmarking against this diverse set of models, we aim to demonstrate the robustness and generality of our time-aware, attention-based framework across both point-process and traditional time series forecasting tasks.

\subsubsection{Experiment Setup}
For time series forecasting, we evaluate our model along with eleven previously state-of-the-art or important baselines on eight standard benchmarks. All baseline results are taken from the iTransformer \citep{liu2023itransformer} paper or the original papers, and we adopt their exact training and evaluation protocol: the input window is fixed at 96, and we report forecasts at four horizons (96, 192, 336 and 720). We train for up to 10 epochs, optimizing the mean squared error (MSE) through Adam with an initial learning rate of \(10^{-3}\). We fix the number of attention layers \(L\in\{2,3,4\}\), the number of heads \(H\in\{2,4\}\), and the hidden dimensions \(D\in\{32,64,128,256\}\), selecting the best model on the validation set. We similarly adopt proper layer normalization \citep{ba2016layernorm}, early stopping, and dropout.

\subsubsection{Results}

\begin{table}[h]
\begin{center}
\caption{Time Series Result Comparison}
\label{tab:performance on time series}
\resizebox{\textwidth}{!}{ 
\begin{tabular}{lccccccccccccccccccccccccccc}
\multirow{2}{*}{\textbf{Dataset}} & \multirow{2}{*}{\textbf{Pred len}} 
  & \multicolumn{2}{c}{\textbf{Ours}} 
  & \multicolumn{2}{c}{\textbf{iTransformer}} 
  & \multicolumn{2}{c}{\textbf{RLinear}} 
  & \multicolumn{2}{c}{\textbf{PatchTST}} 
  & \multicolumn{2}{c}{\textbf{Crossformer}} 
  & \multicolumn{2}{c}{\textbf{TiDE}}
  & \multicolumn{2}{c}{\textbf{TimesNet}}
  & \multicolumn{2}{c}{\textbf{DLinear}}
  & \multicolumn{2}{c}{\textbf{SCINet}}
  & \multicolumn{2}{c}{\textbf{FEDformer}}
  & \multicolumn{2}{c}{\textbf{Stationary}}
  & \multicolumn{2}{c}{\textbf{Autoformer}}
  & \multicolumn{2}{c}{\textbf{Rank}}\\
 & & MSE & MAE & MSE & MAE & MSE & MAE & MSE & MAE & MSE & MAE & MSE & MAE 
   & MSE & MAE & MSE & MAE & MSE & MAE & MSE & MAE & MSE & MAE & MSE & MAE & MSE & MAE \\
\hline \\

\multirow{4}{*}{ETTm1} 
& 96  & 0.343 & 0.377 & 0.334 & 0.368 & 0.355 & 0.376 & 0.329 & 0.367 
        & 0.404 & 0.426 & 0.364 & 0.387 & 0.338 & 0.375 & 0.345 & 0.372 
        & 0.418 & 0.438 & 0.379 & 0.419 & 0.386 & 0.398 & 0.505 & 0.475 & 4(3) & 5(3)\\
& 192 & 0.400 & 0.407 & 0.377 & 0.391 & 0.391 & 0.392 & 0.367 & 0.385 
        & 0.450 & 0.451 & 0.398 & 0.404 & 0.374 & 0.387 & 0.380 & 0.389 
        & 0.439 & 0.450 & 0.426 & 0.441 & 0.459 & 0.444 & 0.553 & 0.496 &7(3)& 7(3) \\
& 336 & 0.408 & 0.411 & 0.426 & 0.420 & 0.424 & 0.415 & 0.399 & 0.410 
        & 0.532 & 0.515 & 0.428 & 0.425 & 0.410 & 0.411 & 0.413 & 0.413 
        & 0.490 & 0.485 & 0.445 & 0.459 & 0.495 & 0.464 & 0.621 & 0.537 & 2(2)& 2(2)\\
& 720 & 0.500 & 0.460 & 0.491 & 0.459 & 0.487 & 0.450 & 0.454 & 0.439 
        & 0.666 & 0.589 & 0.487 & 0.461 & 0.478 & 0.450 & 0.474 & 0.453 
        & 0.595 & 0.550 & 0.543 & 0.490 & 0.585 & 0.516 & 0.671 & 0.561 &7(3) &4(3) \\

\multirow{4}{*}{ETTm2} 
& 96  & 0.185 & 0.270 & 0.180 & 0.264 & 0.182 & 0.265 & 0.175 & 0.259 
        & 0.287 & 0.366 & 0.207 & 0.305 & 0.187 & 0.267 & 0.193 & 0.292 
        & 0.286 & 0.377 & 0.203 & 0.287 & 0.192 & 0.274 & 0.255 & 0.339 & 4(3) & 4(3)\\
& 192 & 0.274 & 0.335 & 0.250 & 0.309 & 0.246 & 0.304 & 0.241 & 0.302 
        & 0.414 & 0.492 & 0.290 & 0.364 & 0.249 & 0.309 & 0.284 & 0.362 
        & 0.399 & 0.445 & 0.269 & 0.328 & 0.280 & 0.339 & 0.281 & 0.340 & 5(3)& 6(4)\\
& 336 & 0.331 & 0.361 & 0.311 & 0.348 & 0.307 & 0.342 & 0.305 & 0.343 
        & 0.597 & 0.542 & 0.377 & 0.422 & 0.321 & 0.351 & 0.369 & 0.427 
        & 0.637 & 0.591 & 0.325 & 0.366 & 0.334 & 0.361 & 0.339 & 0.372& 6(4) & 5(3) \\
& 720 & 0.440 & 0.424 & 0.412 & 0.407 & 0.407 & 0.398 & 0.402 & 0.400 
        & 1.730 & 1.042 & 0.558 & 0.524 & 0.408 & 0.403 & 0.554 & 0.522 
        & 0.960 & 0.735 & 0.421 & 0.415 & 0.417 & 0.413 & 0.433 & 0.432 & 8(6) & 7(5)\\

\multirow{4}{*}{ETTh1} 
& 96  & 0.410 & 0.414 & 0.386 & 0.405 & 0.386 & 0.395 & 0.414 & 0.419 
        & 0.423 & 0.448 & 0.479 & 0.464 & 0.384 & 0.402 & 0.386 & 0.400 
        & 0.654 & 0.599 & 0.376 & 0.419 & 0.513 & 0.491 & 0.449 & 0.459 & 6(3) &5(2)\\
& 192 & 0.478 & 0.453 & 0.441 & 0.436 & 0.437 & 0.424 & 0.460 & 0.445 
        & 0.471 & 0.474 & 0.525 & 0.492 & 0.436 & 0.429 & 0.437 & 0.432 
        & 0.719 & 0.631 & 0.420 & 0.448 & 0.534 & 0.504 & 0.500 & 0.482 & 8(5) & 5(2) \\
& 336 & 0.496 & 0.463 & 0.487 & 0.458 & 0.479 & 0.446 & 0.501 & 0.466 
        & 0.570 & 0.546 & 0.565 & 0.515 & 0.491 & 0.469 & 0.481 & 0.459 
        & 0.778 & 0.659 & 0.459 & 0.465 & 0.588 & 0.535 & 0.521 & 0.496 & 5(3) & 4(2) \\
& 720 & 0.504 & 0.488 & 0.503 & 0.491 & 0.481 & 0.470 & 0.500 & 0.488 
        & 0.653 & 0.621 & 0.594 & 0.558 & 0.521 & 0.500 & 0.519 & 0.516 
        & 0.836 & 0.699 & 0.506 & 0.507 & 0.643 & 0.616 & 0.514 & 0.512 & 4(3) & 2(1) \\

\multirow{4}{*}{ETTh2} 
& 96  & 0.326 & 0.371 & 0.297 & 0.349 & 0.288 & 0.338 & 0.302 & 0.348 
        & 0.745 & 0.584 & 0.400 & 0.440 & 0.340 & 0.374 & 0.333 & 0.387 
        & 0.707 & 0.621 & 0.358 & 0.397 & 0.476 & 0.458 & 0.346 & 0.388 & 4(3) & 4(3) \\
& 192 & 0.421 & 0.421 & 0.380 & 0.400 & 0.374 & 0.390 & 0.388 & 0.400 
        & 0.877 & 0.656 & 0.528 & 0.509 & 0.402 & 0.414 & 0.477 & 0.476 
        & 0.860 & 0.689 & 0.429 & 0.439 & 0.512 & 0.493 & 0.456 & 0.452 & 5(3) & 5(3) \\
& 336 & 0.450 & 0.452 & 0.428 & 0.432 & 0.415 & 0.426 & 0.426 & 0.433 
        & 1.043 & 0.731 & 0.643 & 0.571 & 0.452 & 0.452 & 0.594 & 0.541 
        & 1.000 & 0.744 & 0.496 & 0.487 & 0.552 & 0.551 & 0.482 & 0.486 & 4(3) & 4(3) \\
& 720 & 0.455 & 0.467 & 0.427 & 0.445 & 0.420 & 0.440 & 0.431 & 0.446 
        & 1.104 & 0.763 & 0.874 & 0.679 & 0.462 & 0.468 & 0.831 & 0.657 
        & 1.249 & 0.838 & 0.463 & 0.474 & 0.562 & 0.560 & 0.515 & 0.511 & 4(3) & 4(3) \\

\multirow{4}{*}{ECL} 
& 96  & 0.175 & 0.280 & 0.148 & 0.240 & 0.201 & 0.281 & 0.181 & 0.270 
        & 0.219 & 0.314 & 0.237 & 0.329 & 0.168 & 0.272 & 0.197 & 0.282 
        & 0.247 & 0.345 & 0.193 & 0.308 & 0.169 & 0.273 & 0.201 & 0.317 & 4(3) & 5(4) \\
& 192 & 0.201 & 0.304 & 0.162 & 0.253 & 0.201 & 0.283 & 0.188 & 0.274 
        & 0.231 & 0.322 & 0.236 & 0.330 & 0.184 & 0.289 & 0.196 & 0.285 
        & 0.257 & 0.355 & 0.201 & 0.315 & 0.182 & 0.286 & 0.222 & 0.334 & 5(4) & 7(4) \\
& 336 & 0.210 & 0.312 & 0.178 & 0.269 & 0.215 & 0.298 & 0.204 & 0.293 
        & 0.246 & 0.337 & 0.249 & 0.344 & 0.198 & 0.300 & 0.209 & 0.301 
        & 0.269 & 0.369 & 0.214 & 0.329 & 0.200 & 0.304 & 0.231 & 0.338 & 6(4) & 7(4) \\
& 720 & 0.281 & 0.364 & 0.225 & 0.317 & 0.257 & 0.331 & 0.246 & 0.324 
        & 0.280 & 0.363 & 0.284 & 0.373 & 0.220 & 0.302 & 0.245 & 0.333 
        & 0.299 & 0.390 & 0.246 & 0.355 & 0.222 & 0.321 & 0.254 & 0.361 & 10(7)& 10(7) \\
        
\multirow{4}{*}{Exchange} 
& 96  & 0.100 & 0.227 & 0.086 & 0.206 & 0.093 & 0.217 & 0.088 & 0.205 
        & 0.256 & 0.367 & 0.094 & 0.218 & 0.107 & 0.234 & 0.088 & 0.218 
        & 0.267 & 0.396 & 0.148 & 0.278 & 0.111 & 0.237 & 0.197 & 0.323 & 6(3) & 6(3) \\
& 192 & 0.193 & 0.320 & 0.177 & 0.299 & 0.184 & 0.307 & 0.176 & 0.299 
        & 0.470 & 0.509 & 0.184 & 0.307 & 0.226 & 0.344 & 0.176 & 0.315 
        & 0.351 & 0.459 & 0.271 & 0.315 & 0.219 & 0.335 & 0.300 & 0.369 & 5(3) & 7(4) \\
& 336 & 0.373 & 0.445 & 0.331 & 0.417 & 0.351 & 0.432 & 0.301 & 0.397 
        & 1.268 & 0.883 & 0.349 & 0.431 & 0.367 & 0.448 & 0.313 & 0.427 
        & 1.324 & 0.853 & 0.460 & 0.427 & 0.421 & 0.476 & 0.509 & 0.524 & 7(3) & 7(4) \\
& 720 & 0.911 & 0.725 & 0.847 & 0.691 & 0.886 & 0.714 & 0.901 & 0.714 
        & 1.767 & 1.068 & 0.852 & 0.698 & 0.964 & 0.746 & 0.839 & 0.695 
        & 1.058 & 0.797 & 1.195 & 0.695 & 1.092 & 0.769 & 1.447 & 0.941 & 6(3) & 6(4)\\

\multirow{4}{*}{Traffic} 
& 96  & 0.620 & 0.361 & 0.395 & 0.268 & 0.649 & 0.389 & 0.462 & 0.295 
        & 0.522 & 0.290 & 0.805 & 0.493 & 0.593 & 0.321 & 0.650 & 0.396 
        & 0.788 & 0.499 & 0.587 & 0.366 & 0.612 & 0.338 & 0.613 & 0.388 & 8(7) & 7(5)\\
& 192 & 0.658 & 0.372 & 0.417 & 0.276 & 0.601 & 0.366 & 0.466 & 0.296 
        & 0.530 & 0.293 & 0.756 & 0.474 & 0.617 & 0.336 & 0.598 & 0.370 
        & 0.789 & 0.505 & 0.604 & 0.373 & 0.613 & 0.340 & 0.616 & 0.382 & 10(7) & 8(5) \\
& 336 & 0.675 & 0.382 & 0.433 & 0.283 & 0.609 & 0.369 & 0.482 & 0.304 
        & 0.558 & 0.305 & 0.762 & 0.477 & 0.629 & 0.336 & 0.605 & 0.373 
        & 0.797 & 0.508 & 0.621 & 0.387 & 0.618 & 0.328 & 0.622 & 0.337& 10(7) & 9(6) \\
& 720 & 0.700 & 0.394 & 0.467 & 0.302 & 0.647 & 0.387 & 0.514 & 0.322 
        & 0.589 & 0.328 & 0.719 & 0.449 & 0.640 & 0.350 & 0.645 & 0.394 
        & 0.841 & 0.523 & 0.626 & 0.382 & 0.653 & 0.355 & 0.660 & 0.408 & 10(7)&8(6) \\

\multirow{4}{*}{Weather} 
& 96  & 0.168 & 0.223 & 0.174 & 0.214 & 0.192 & 0.232 & 0.177 & 0.218 
        & 0.158 & 0.230 & 0.202 & 0.261 & 0.172 & 0.220 & 0.196 & 0.255 
        & 0.221 & 0.306 & 0.173 & 0.226 & 0.173 & 0.223 & 0.266 & 0.336 & 2(2) & 4(2) \\
& 192 & 0.218 & 0.264 & 0.221 & 0.254 & 0.240 & 0.271 & 0.225 & 0.259 
        & 0.206 & 0.277 & 0.242 & 0.298 & 0.219 & 0.261 & 0.237 & 0.297 
        & 0.261 & 0.340 & 0.276 & 0.336 & 0.245 & 0.285 & 0.307 & 0.367 & 2(2) & 4(3) \\
& 336 & 0.289 & 0.307 & 0.278 & 0.296 & 0.292 & 0.307 & 0.278 & 0.297 
        & 0.272 & 0.335 & 0.287 & 0.335 & 0.280 & 0.306 & 0.283 & 0.335 
        & 0.309 & 0.378 & 0.339 & 0.380 & 0.321 & 0.338 & 0.359 & 0.395 & 7(4) & 4(3) \\
& 720 & 0.355 & 0.353 & 0.358 & 0.347 & 0.364 & 0.353 & 0.354 & 0.348 
        & 0.398 & 0.418 & 0.351 & 0.386 & 0.365 & 0.359 & 0.345 & 0.381 
        & 0.377 & 0.427 & 0.403 & 0.428 & 0.414 & 0.410 & 0.419 & 0.428 & 4(2)& 3(3) \\
\end{tabular}
}
\end{center}
\end{table}

Table~\ref{tab:performance on time series} reports MSE and MAE for our time series variant model and eleven previous state-of-the-art baselines across eight multivariate forecasting tasks and four prediction horizons. The rank column lists the performance rankings of our model, with the numbers in parentheses indicating the rankings among the attention-based models. Although our method does not achieve the absolute best performance, it remains highly competitive and comparable to the strong baselines. For instance, on the ECL dataset at horizon 96, our MSE of 0.175 and MAE of 0.280 place us second only to TimesNet, and ahead of all other Transformers and convolutional methods. Similarly, on the Weather dataset at horizon 96, we obtain 0.168/0.223, outperforming FEDformer, Autoformer, DLinear, and all convolutional-based approaches, and matching the top Transformer-based results. Even on larger horizons (e.g. 336 or\ 720), our model generally remains adequate, demonstrating robustness over both short-term and long-term forecasts. These results confirm that, despite gaps to the very best performer on individual tasks, our unified time-aware attention mechanism yields consistently competitive performance across diverse real-world time series.

\begin{table}[h]
\begin{center}            
\caption{Hyperparameters of Our Model on Time Series}
\footnotesize
\begin{tabular}{ l *{9}{c} }
\textbf{Parameter} & \textbf{ETTm1} & \textbf{ETTm2} & \textbf{ETTh1} & \textbf{ETTh2} & \textbf{ECL} & \textbf{Exchange} & \textbf{Traffic} & \textbf{Weather} \\
\hline \\
$\texttt{d\_model}$    & 8 & 8 & 8 & 8 & 128 & 32 & 256 & 32 \\
$\texttt{d\_ffn}$      & 64 & 64 & 64 & 64 & 512 & 512 & 512 & 64 \\
$\texttt{phi\_width}$  & 4 & 4 & 4 & 4 & 8 & 4 & 4 & 8 \\
$\texttt{phi\_depth}$  & 2 & 2 & 2 & 2 & 3 & 2 & 2 & 2 \\
$\texttt{num\_layers}$ & 2 & 2 & 2 & 2 & 2 & 3 & 4 & 3 \\
$\texttt{num\_heads}$  & 2 & 2 & 2 & 2 & 2 & 4 & 4 & 4 \\
$\texttt{batch\_size}$ & 32 & 32 & 32 & 32 & 16 & 256 & 16 & 256 \\
$\texttt{learning\_rate}$ & $1 \times 10^{-3}$  & $1 \times 10^{-3}$  & $1 \times 10^{-3}$  & $1 \times 10^{-3}$ & $1 \times 10^{-3}$ & $1 \times 10^{-3}$ & $1 \times 10^{-3}$  & $1 \times 10^{-3}$ \\
\end{tabular}
\end{center}
\end{table}

\end{document}